\documentclass[journal]{IEEEtran}

\ifCLASSINFOpdf
\else
   \usepackage[dvips]{graphicx}
\fi
\usepackage{url}
\usepackage{cite}
\usepackage{graphicx}
\usepackage{amssymb}
\usepackage{amsmath}
\usepackage{booktabs}
\usepackage{color}
\usepackage[pagebackref=true,breaklinks=true,citecolor=blue,colorlinks,bookmarks=false]{hyperref}

\begin{document}

%\title{Building Spatial-temporal Context Exchange among Head, Face, and Eye: An End-to-end Video Gaze Estimation Solution}

\title{End-to-end Video Gaze Estimation via Capturing Head-face-eye Spatial-temporal Interaction Context }

\author{Yiran Guan$^*$, Zhuoguang Chen$^*$, Wenzheng Zeng\dag, Zhiguo Cao~\IEEEmembership{Member,~IEEE} and Yang Xiao\dag  
\thanks{This work is supported by the National Natural Science Foundation of China (Grant No. 62271221). }
\thanks{Yiran Guan, Zhuoguang Chen, Wenzheng Zeng, Zhiguo Cao, and Yang Xiao are with National Key Laboratory of Science and Technology on Multi-Spectral Information Processing, School of Artificial Intelligence and Automation, Huazhong University of Science and Technology, China. E-mail: yiranguan, zgchen33, wenzhengzeng, zgcao, Yang\_Xiao@hust.edu.cn. }
\thanks{$*$ Yiran Guan and Zhuoguang Chen are of equal contribution.}
\thanks{\dag~Wenzheng Zeng and Yang Xiao are corresponding authors.}
}

\maketitle
\begin{abstract}
 In this letter, we propose a new method, Multi-Clue Gaze (MCGaze), to facilitate video gaze estimation via capturing spatial-temporal interaction context among head, face, and eye in an end-to-end learning way, which has not been well concerned yet. The main advantage of MCGaze is that the tasks of clue localization of head, face, and eye can be solved jointly for gaze estimation in a one-step way, with joint optimization to seek optimal performance. During this, spatial-temporal context exchange happens among the clues on the head, face, and eye. Accordingly, the final gazes obtained by fusing features from various queries can be aware of global clues from heads and faces, and local clues from eyes simultaneously, which essentially leverages performance. Meanwhile, the one-step running way also ensures high running efficiency. Experiments on the challenging Gaze360 dataset verify the superiority of our proposition. The source code will be released at \url{https://github.com/zgchen33/MCGaze}.

\end{abstract}

\begin{IEEEkeywords}
gaze estimation, video, head-face-eye spatial-temporal context, query%, query method  
\end{IEEEkeywords}

\IEEEpeerreviewmaketitle

\section{Introduction}
\IEEEPARstart{V}{ideo} gaze estimation is a recently emerged challenging research task that suffers from the critical issues of the variations on the pose, human attribute, illumination, etc.
% delete reference: gaze_cognition TIP
It can be widely used to understand human cognitive patterns~\cite{henderson2003human, SPL_gaze_object_segment}, 
% mental states  
% delete reference: video_quality_eye TSP
human social interaction~\cite{fan2018inferring, fan2019understanding, emery2000eyes}, 
% delete reference: TIP_gaze_pattern
and human-machine interaction~\cite{zhang2019evaluation}. Compared with estimating gaze in individual images~\cite{zhang2015appearance}, richer spatial-temporal context over head, face, and eye is essentially involved in video setting, which is beneficial for better characterizing gaze patterns.
% delete reference: Krafka_Khosla_Kellnhofer_Kannan_Bhandarkar_Matusik_Torralba_2016
Although the paid efforts~\cite{2eye_gaze,nonaka2022dynamic,bao2021adaptive,cheng2020coarse,tip_gaze}, we argue that they still have not well captured the spatial-temporal descriptive clues as below: 

$\bullet$ First of all, the interaction among head, face, and eye features has not been established, for distilling the underly video gaze characterization context;

$\bullet$ Secondly, tasks of gaze estimation, and clue localization of head, face, and eye cannot be jointly solved with joint optimization to seek optimal performance; 

$\bullet$ Last but not least, multi-clue spatial and continuous temporal features cannot be extracted holistically within a unified framework.

\begin{figure}[t]
\centerline{\includegraphics[width=\columnwidth]{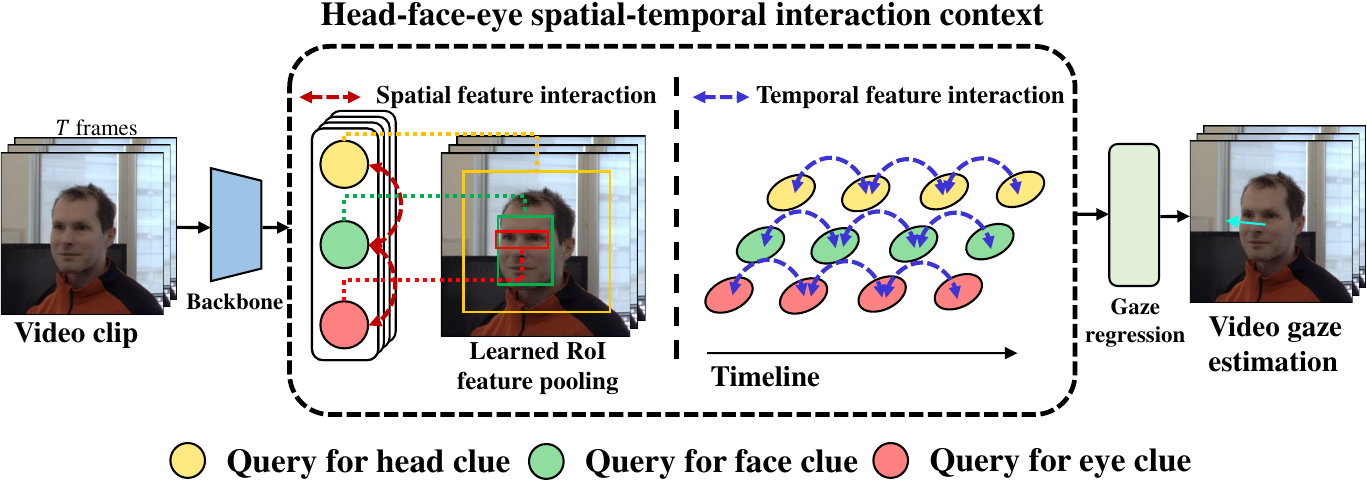}}
\caption{The main idea of MCGaze. It facilitates gaze estimation performance via concerning head-face-eye spatial-temporal interaction context with multi-clue feature fusion.}
\label{fig:main_idea}
\vspace{-4mm}
\end{figure}

To address these, we propose MCGaze, a video gaze estimation method that facilitates performance by capturing the head-face-eye spatial-temporal interaction context in an end-to-end query-based learning way. Meanwhile, the tasks of gaze estimation and clue localization of the head, face, and eye can be solved integrally in a one-step running way.%with joint optimization.

Particularly, our main idea is shown in Fig.~\ref{fig:main_idea}. Towards a gaze video clip, its per-frame features will be first extracted to form a video feature tensor. Then, the learnable queries of spatial-temporal form on the head, face, and eye will be set up to take the roles of localizing clues on the head, face, and eye for gaze characterization jointly. At each time point, the frame-wise feature interaction among head, face, and eye queries is executed via spatial interaction for information exchange between the global descriptive clues on head and face, and the local fine clues on eyes. Accordingly, each type of query will be of strong local-global gaze characterization ability. More specifically, head and face clues can reveal human pose, human attributes, and illumination information. And, eye clues essentially characterize the gaze's fine details. On the other hand, within each query, feature interaction between neighboring frames via temporal interaction is also performed to capture the motion information on the head, face, and eye to leverage sequential gaze estimation and facilitate temporal consistency. Finally, features from the head, face, and eye will be jointly used for gaze estimation.
\begin{figure*}[t]
\vspace{-6mm}
\begin{center}
\includegraphics[width=0.85\textwidth]{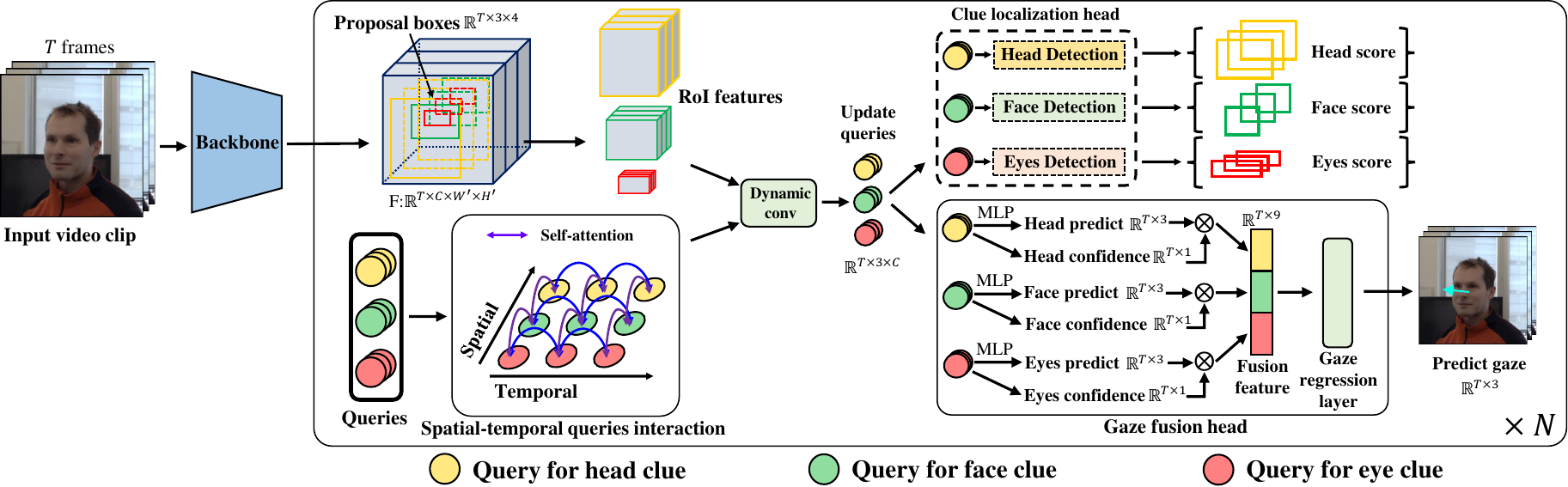}
\vspace{-5mm}
\end{center}
        \caption{The main technical pipeline of MCGaze.}
\label{fig:pipeline}
\vspace{-5mm}
\end{figure*}
% delete reference: face_d
It is worth noting that, the procedures of gaze estimation, and clue localization of head, face, and eye are conducted in a one-step running way, with joint optimization to seek the optimal performance. That is to say, unlike previous works, we do not need to use a face detector~\cite{mtcnn} or eye detector~\cite{SPL_Facial_Landmark,faciallandmark_tip_23} to preprocess the input head images. This manner can help ensure high running efficiency due to feature sharing among the tasks, which practical applications prefer. The experiments on the challenging Gaze360 dataset~\cite{kellnhofer2019gaze360} verify the superiority of our proposition for video gaze estimation.

Overall, our main contributions can be summarized as:

$\bullet$ A novel end-to-end video gaze estimation method is proposed, via capturing head-face-eye spatial-temporal interaction context to facilitate performance;

$\bullet$ Video gaze estimation, and clue localization of head, face, and eye can be solved integrally in a one-step running way, with joint optimization to seek optimal performance.

\section{APPROACH}
% \begin{figure*}[t]
% \begin{center}

% \includegraphics[width=0.9\textwidth]{pipeline.png}
% \vspace{-5mm}
% \end{center}
%         \caption{Overview of our gaze estimation method.}
% \label{fig:pipeline}
% \vspace{-5mm}
% \end{figure*}

\subsection{Overall Method}  
 In this section, we present our proposed method, MCGaze. Taking a video clip as input, it can automatically capture head, face, and eye clues for hierarchical spatial-temporal gaze representation, and predict the gaze direction of each frame in the video. 
Our method employs spatial-temporal interactions among head-face-eye clues throughout the video clip.  It draws inspiration from query-based methods~\cite{sparse,Tevit,instblink,garg2023multiscaled,spl_query_specific,2023query_video} and local-global spatial-temporal modeling approaches~\cite{local2023,local2020,liang2023local,cycmunet2023}.  
 % Our design is inspired by query-based methods  \cite{sparse,Tevit,instblink,garg2023multiscaled,spl_query_specific,2023query_video}, and leveraged by spatial-temporal interactions among head-face-eye clues across the entire video clip. 
 The architecture is illustrated in Fig.~\ref{fig:pipeline}.
 
 % Given a video clip $I\in \mathbb{R}^{T\times 3 \times H \times W}$, where $T$ represents the number of frames and $3 \times H \times W$ represents each input frame as an RGB image of size $H \times W$, our method applies a backbone network to extract features from the video input, resulting in $F \in \mathbb{R}^{T\times C \times H^{'} \times W^{'}}$, where C denotes the number of channels and $H^{'}\times W^{'}$ represents the size of the feature maps.
 Specifically, our method applies a backbone network to extract features from a video clip $I\in \mathbb{R}^{T\times 3 \times H \times W}$. Here, $T$ represents the number of frames, and $3 \times H \times W$ represents the input frame as an RGB image of size $H \times W$. Then, the backbone network generates $F \in \mathbb{R}^{T\times C \times H^{'} \times W^{'}}$, where $C$ represents the number of channels and $H^{'}\times W^{'}$ denotes the size of the feature maps.
% Next, the extracted features are fed into our clue-based architecture, which iterates N times and consists of two main components: the Clue Interaction Module and Task-specific Heads (i.e., \textbf{queries anchor head} and gaze fusion head). In each iteration, queries are updated and the \textbf{queries anchor head} produces outputs for head, face, eye positions and classification scores, and gaze fusion head produces gaze direction. The output of the final iteration serves as the ultimate result.

Next, the extracted features are fed into our query-based architecture, which iterates $N$ times and consists of two main components: the spatial-temporal query interaction and the task-specific heads (i.e., clue localization head and gaze fusion head). In each iteration, the queries for the head, face, and eye clue are updated, and the clue localization head predicts the clue region of the head, face, and eye. On the other hand, the gaze fusion head determines the direction of the human gaze from the head, face, and eye clue. The gaze predicted by the last iteration is used as the output of the model.

\subsection{Head-face-eye Queries}
 % In our method, multi-level queries ${q_{clue}}$, where $clue \in \{ head, face, eye \}$ are responsible for describing specific regions of interest for the subject's head, face, and eyes in each frame of the video. Each clue consists of T embeddings with a feature dimension of C, denoted as $q_{clue}\in \mathbb{R}^{T \times C}$. Each embedding focuses on the features within the corresponding region of interest in the frame. Additionally, there are proposal boxes $p_{clue}\in \mathbb{R}^{T \times 4}$ that correspond to each clue. The proposal boxes for each frame indicate the positions of the head, face, and eyes in the feature map. Both $p_{clue}$ and $q_{clue}$ are learnable and update their parameters in each iteration to achieve feature localization and extraction for the target regions.
 % In our method, multi-level clues $clue_{i}$, where $i \in \{ head, face, eye \}$, are responsible for describing specific regions of interest for the subject's head, face, eyes and gaze representation in each frame of the video. Each clue consists of T embeddings with a feature dimension of C, denoted as $clue_{i} \in \mathbb{R}^{T \times C}$. Additionally, there are proposal boxes $p_{i}\in \mathbb{R}^{T \times 4}$ that correspond to each clue. The proposal boxes for each frame indicate the positions of the head, face, and eyes in the feature map. Both $clue_{i}$ and $p_{i}$ are learnable and their parameters are updated in each iteration to achieve extraction for the target regions and gaze representation.
 Our approach applies multi-clue queries $q_{clue} \in \mathbb{R}^{T \times C}$, $clue \in \{head, face, eye\}$ to capture the subject's corresponding clue regions and gaze representations from it in the video. Each query comprises $T$ embeddings with a feature dimension of $C$. Each embedding generally focuses on the feature representation of the corresponding frame. Additionally, corresponding to each query, there exist proposal boxes $p_{clue}\in \mathbb{R}^{T \times 4}$ that indicate the locations of the subject's head, face, and eye in the feature map. The parameters of both $q_{clue}$ and $p_{clue}$ are learnable. For each complete forward propagation, they will be updated in an iterative way to achieve effective extraction of target clues and gaze representations from it.
 
\subsection{Spatial-temporal Queries Interaction (STQI)}
Local-global spatial-temporal modeling is very important for the video task\cite{local2023,local2020,liang2023local,cycmunet2023}, here we design specific queries for the three key clues for our task. Inspired by the transformer structure, we build strong interaction among spatial and temporal dimensions to facilitate gaze representation. Specifically, we use spatial-temporal queries interaction module~\cite{Tevit} to better localize the hierarchical clues and build effective information exchange for robust gaze representations. In this module, a spatial self-attention layer is used to enable spatial interaction among head, face, and eye query within the same frame:
\begin{equation}
\{q_{head}^{t},q_{face}^{t},q_{eye}^{t}\} = \operatorname{MHSA}(\{q_{head}^{t},q_{face}^{t},q_{eye}^{t}\}),
% \{q_{clue}^{t}\}_{clue \in W} =MHSA(\{q_{clue}^{t}\}_{clue \in W}), 
\end{equation}
% The abbreviation MHSA stands for multi-head self-attention \cite{attention}.For $t\ $in $\left[0,T-1\right]$.These three types of clues facilitate strong interactions at different granularities (head, face, and eye) in the spatial domain.
where $t \in \left[0,T-1\right]$, and the abbreviation $\operatorname{MHSA}$ stands for multi-head self-attention \cite{attention}. Actually, these three types of queries with $\operatorname{MHSA}$ can essentially promote the information exchange among the head and face of global clues and the eye of local clues within the spatial domain. This leads the queries to be of both global and local spatial perspectives for gaze characterization.  
% These three types of query for the corresponding clue strongly promote interactions between different levels of detail, namely the head, face, and eye, within the spatial domain.

Moreover, we apply a self-attention layer to enable temporal interaction for each query along the temporal dimension:
\begin{equation}
\{q_{clue}^{t}\}_{t=1}^{T} = \operatorname{MHSA}(\{q_{clue}^{t}\}_{t=1}^{T}),
% \{clue_{i}^{t}\}_{t=0}^{T-1} = MHSA(\{clue_{i}^{t}\}_{t=0}^{T-1}) 
\end{equation}
where $clue \in \{ head, face, eye \}$. Applying temporal interaction on each query promotes sequential modeling of distinctive features, such as pose variation and eye movement, and facilitates temporal consistency, leading to robust clue localization and gaze estimation.

% After the self-attention layers, aiming at collecting the clues and gaze information, we incorporate a dynamic convolutional layer whose inputs are the learned queries for clues and their corresponding RoI features. 
To let the query acquire highly relevant features from input video features, we use dynamic convolution~\cite{sparse} acting on an RoI feature to update the query's features within each iteration. Specifically, the RoI feature is obtained by RoI align~\cite{maskrcnn} based on the proposal boxes $p_{clue}$. The output feature from dynamic convolution will be used to update query features. The updated query feature $q_{clue}^{*}$ will be used to perform clue localization and gaze estimation by task-specific heads.

\subsection{Task-specific Heads}
% 2 task-specific heads (i.e., clue localization and gaze fusion head) are proposed for clue localization and gaze estimation.
We design two task-specific heads (i.e., clue localization and gaze fusion head) for clue localization and gaze estimation.

% clue localization head and the gaze fusion head. 
% The clue localization head aims for clue localization, while the gaze fusion head aims to fuse the gaze representation from make clues to predict the final gaze direction prediction.

\subsubsection{\textbf{Clue localization head}}

% Given an updated clue feature, we can obtain both the region of interest for the clue and the confidence level of the region content. For clues $clue_{i}$ with different clues, we use three Multilayer Perceptions (MLP) with the same structure but with mutually exclusive parameters, and a sigmoid function to normalize the output of the MLP to represent the confidence level $s_{clue}\in \mathbb{R}^{T}$ for the different clues.
Given an updated query, we can obtain the corresponding clue region that the query focuses on by the clue localization head. For each query $q^{*}_{clue}$, we use a multilayer Perception ($\operatorname{MLP}$) followed by a sigmoid normalization to indicate the clue region existence (e.g., the face or eye cannot be detected when the subject's head is turned back to the camera) $s_{clue}\in \mathbb{R}^{T}$ for the different $clue \in \{head, face, eye\}$:
\begin{equation}
    s_{clue}= \operatorname{Sigmoid}(\operatorname{MLP}^{s}_{clue}(q^{*}_{clue})).
\end{equation}

Similarly, we employ three separate multilayer perceptions to accomplish clue region localization for $clue \in \{head, face, eye\}$:
% anchor the regions of interest for different clues and output the bounding box coordinates $bbox_{clue}\in \mathbb{R}^{T\times4}$ of the head, face and eye region:
\begin{equation}
    b_{clue}= \operatorname{MLP}^{b}_{clue}(q^{*}_{clue}),
\end{equation}
where $b_{clue}$ indicates the clue region localization and will be used to update the proposal boxes $p_{clue}$.

\subsubsection{\textbf{Gaze fusion head}}
For the updated query features of the three clues $q^{*}_{head}\text{, }q^{*}_{face}\text{ and }q^{*}_{eye}$, we use three different $\operatorname{MLP}$s to regress the gaze vectors $g_{clue}$ from them as
\begin{equation}
g_{clue}=\operatorname{MLP}^g_{clue}(q^{*}_{clue}),
\end{equation}

where $g_{clue}\in \mathbb{R}^{3}$ and $clue \in \{ head, face, eye \}$. In fact, the reliability of the gaze prediction obtained from different clues may vary in different situations. For instance, when the head is turned backward, the eyes are not visible, resulting in a low reliability of gaze prediction using the eye clue. Therefore, We use three MLPs to predict the confidence level $c_{clue}$ of the three predicted gazes as
\begin{equation}
    c_{clue}=\operatorname{MLP}^c_{clue}(q^{*}_{clue}).
\end{equation}
Then we multiply the gaze vectors from different queries by their corresponding confidence and concatenate the resulting products.
The final gaze direction $g_{fusion}\in \mathbb{R}^{T\times 3}$ after fusion is output by a fully connected ($\operatorname{FC}$) layer as
\begin{equation}
    g_{fusion}=\operatorname{FC}([g_{head}\times c_{head},g_{face}\times c_{face},g_{eye}\times c_{eye}]).
\end{equation}
% \begin{equation}
% \begin{split}
% g_{clue}=MLP^g_{clue}(q^{*}_{clue}),\\c_{clue}=MLP^c_{clue}(q^{*}_{clue}),\\
% g_{fusion}=FC([g_{head}\times c_{head},g_{face}\times c_{face},g_{eye}\times c_{eye}]).
% \end{split}
% \end{equation}

\subsection{Model Training}
% To train our network, we designed several loss functions to supervise the output of task-specific heads. For ease of description, we use QAH instead of Queries Anchor Head, and GFH instead of Gaze Fusion Head. We supervise $s_{clue}$ and $b_{clue}$ using $\mathcal{L}_{cls}$ and $\mathcal{L}_{box}$, where $\mathcal{L}_{cls}$ uses focal loss for clue existence classification. $\mathcal{L}_{box}$ is the combination of L1 loss and GIoU loss to locate a specific clue. So the loss of QAH is  formulated as
We design several loss functions to optimize the whole network. In order to have the clues anchor at the target level (i.e., head, face, and eye), we supervise the clue region existence $s_{clue}$ and bounding box location $b_{clue}$ using  $\mathcal{L}_{cls}$ and $\mathcal{L}_{box}$ respectively, where $\mathcal{L}_{cls}$ indicates the focal loss \cite{focalloss}. $\mathcal{L}_{box}$ indicates the combination of L1 loss and GIoU loss \cite{Giou} for bounding box regression. Specifically, the loss  is formulated as
\begin{equation}
    \mathcal{L}_{anchor} = \sum_{t=0}^{T-1}\sum_{clue}(\mathcal{L}_{box}(b^t_{clue},\hat{b}^t_{clue})+\mathcal{L}_{cls}(s^t_{clue},\hat{s}^t_{clue})),
\end{equation}
where $clue \in \{head, face, eye\}$. Besides, we use $\arccos$ loss to supervise gaze estimation, whose expression is
\begin{equation}
    \mathcal{L}_{arccos}=\arccos {\frac{g \cdot \hat{g}}{\Vert g \Vert \Vert \hat{g} \Vert}},
\end{equation} where $\hat{g}$ denotes the output predicted gaze and $g$ denotes the ground-truth gaze. Besides the final output $g_{fusion}$ from the gaze fusion head, we also supervise the gaze prediction result within each individual clue to make them close to the real gaze direction.
Specifically, the loss of gaze estimation is formulated as 
\begin{equation}
    \mathcal{L}_{gaze}=\sum_{t=0}^{T-1}(\mathcal{L}_{arccos}(g_{fusion}^{t},\hat{g}^t)+\sum_{clue}\mathcal{L}_{arccos}(g^t_{clue},\hat{g}^t)),
\end{equation}
where $clue \in \{head, face, eye\}$. In addition, for better temporal modeling and to ensure the temporal stability of the output gaze, we add the temporal regularization term $\mathcal{J}_{temp}$ with the expression:
\begin{equation}
 \mathcal{J}_{temp}=\sum_{t=1}^{T-2}\vert 2\times \hat{g}_{fusion}^{t}-\hat{g}_{fusion}^{t+1}-\hat{g}_{fusion}^{t-1} \vert,  
\end{equation}
where $\hat{g}^{t}$ denotes the t-th frame of the output gaze. our overall loss function is designed as
\begin{equation}
% \mathcal{L}_{total}=\lambda_1 \mathcal{L}_{anchor}+\lambda_2 \mathcal{L}_{gaze}+\lambda_3 \mathcal{J}_{temp},
\mathcal{L}_{total}= \mathcal{L}_{anchor}+\lambda_1 \mathcal{L}_{gaze}+\lambda_2 \mathcal{J}_{temp},
\end{equation}
where$\ \lambda_1,\lambda_2$ represent the hyperparameters in the loss function. In our experiments, they are set to 6 and 1 respectively.
\section{EXPERIMENTS}
\subsection{Dataset}
To verify the superiority and effectiveness of MCGaze, it is tested on the challenging video gaze estimation dataset Gaze360~\cite{kellnhofer2019gaze360}. It involves 238 subjects under indoor and outdoor environments with labeled 3D gaze with variational head poses and imaging distances.
% Subjects in the dataset presented a wide variety of head postures, with almost one-third of the video frames failing to identify the subject's face. 
% Actually, recent work extracted frames in Gaze360 where the subject’s face could be recognized and used them for model evaluation. This is because the images that cannot detect face are not suitable for appearance-based methods. Therefore, In our work, we train and evaluate our model using  sub-dataset of Gaze360 that can detect face, as well as entire Gaze360 dataset.

Recent researches~\cite{cheng2022gaze,abdelrahman2022l2cs,yan2023gaze} conduct evaluation on the face-detectable subset of the Gaze360 dataset. The reason is that some samples within Gaze360 only capture the back side of the subject whose eyes are not visible and thus unsuitable for appearance-based methods. Following the main evaluation procedure of the recent works~\cite{cheng2022gaze,abdelrahman2022l2cs,yan2023gaze}, we train and evaluate our model on the face-detectable sub-dataset of Gaze360 which we refer to as \textbf{the detectable face setting}.
Besides, we also conduct experiments on the entire Gaze360 to compare with some earlier works~\cite{kellnhofer2019gaze360,kothari2021weakly} that focused on all 360 degrees which we refer to as \textbf{the $\boldsymbol{360^\circ}$ setting}.

\textbf{Evaluation metirc.} Following most of works~\cite{kellnhofer2019gaze360,cheng2022gaze,abdelrahman2022l2cs,yan2023gaze, 2eye_gaze}, angular error ($^\circ$) is used to measure the accuracy of 3D gaze estimation, with the following expression: 
\begin{equation}
% MAE=\frac{1}{K}\sum_{i=1}^{K}\arccos {\frac{g_i \cdot \hat{g}_i}{\Vert g_i \Vert \Vert \hat{g}_i \Vert }},  
\mathcal{L}_{angular}={\frac{g \cdot \hat{g}}{\Vert g \Vert \Vert \hat{g} \Vert }},  
\end{equation}
where $\hat g\in\mathbb{R}^{3}$ is the predicted gaze vector; ${g}\in\mathbb{R}^{3}$ is the ground-truth gaze direction. 
% The smaller MAE value indicates that the method has better performance because it expresses the angular error ($ ^\circ$) between the predicted vector and the ground-truth vector.
\subsection{Implementation details}
On the detectable face setting, we use ResNet-50-FPN~\cite{he2016deep,lin2017feature} backbone. 
The ResNet-50 is pre-trained on ImageNet-1K~\cite{deng2009imagenet} and the iteration time $N$ is set to 4. The model is trained using AdamW~\cite{adam} optimizer with a batch size of 8. The initial learning rate is set to 1e-4 for the backbone and 1e-3 for the other components. During training, the input video clip length is set to 7, and before being fed into the network, frames are resized to 448 × 448 following L2CS-Net baseline~\cite{abdelrahman2022l2cs}. We train the model for 13,000 iterations, with the learning rate decreasing by a factor of 0.1 at iteration 12,000. During testing, we set the input video clip length to 7 with a stride of 4 and employ temporal smoothing.
On the $360^\circ$ setting, the experimental details are similar to those on the detectable face setting. The differences are that the frames are resized to 224 × 224 following Gaze360 baseline~\cite{kellnhofer2019gaze360} for a fair comparison, and the batch size is set to 32. 
% All experiments are conducted on a single RTX 3090 and no Test-Time Augmentation is used in our method.
All experiments are conducted on a single RTX 3090 and no Test-Time Augmentation is used in any of our experiments.
% All experiments are conducted on one RTX 3090. 

\begin{table}[t]
\vspace{-6mm}
\setlength{\abovecaptionskip}{0cm}  %段前
\setlength{\belowcaptionskip}{-0.2cm} %段后
\caption{Comparison on sub-dataset of Gaze360\\ that can detect face.}
\centering
\tiny
\setlength{\tabcolsep}{3pt}
\begin{tabular}{c|ccc}
 \hline
Method & Detectable faces & Front$180^{\circ}$ & Front facing\\
 \hline
FullFace~\cite{zhang2017s} & 14.99 & N/A & N/A \\
RT-Gene~\cite{fischer2018rt} & 12.26 & N/A & N/A \\
Dilated-Net~\cite{chen2019appearance} & 13.73 & N/A & N/A \\
Gaze360~\cite{kellnhofer2019gaze360} & 11.04 & N/A & N/A \\
CA-Net~\cite{cheng2020coarse} & 11.20 & N/A & N/A \\
GazeTR~\cite{cheng2022gaze} & 10.62 & N/A & N/A \\
L2CS-Net~\cite{abdelrahman2022l2cs} & 10.60 & 10.41 & 9.04\\
SPMCCA-Net~\cite{yan2023gaze} & N/A & 10.13 & 8.40\\
CADSE~\cite{CADSE} & 10.70 & N/A & N/A\\
GazeNAS-ETH~\cite{NASgaze2023} & 10.52 & N/A & N/A\\
 \hline
MCGaze~(Ours) & \textbf{10.02} & \textbf{9.81} & \textbf{7.57} \\
 \hline
\end{tabular}
\label{tab: sub-dataset1}
\vspace{-4mm}
\end{table}

\begin{table}[t]
\setlength{\abovecaptionskip}{0cm}  %段前
\setlength{\belowcaptionskip}{-0.2cm} %段后
\caption{Comparison on the entire Gaze360 dataset.}
\centering
\tiny
\setlength{\tabcolsep}{3pt}
\begin{tabular}{c|ccc}
 \hline
Method & All$360^{\circ}$ & Front$180^{\circ}$ & Front facing \\ 
 \hline
Gaze360~\cite{kellnhofer2019gaze360} & 13.50 & 11.40 & 11.10\\
LEAO~\cite{kothari2021weakly} & 13.20 & N/A & 10.10\\
% Bot2L-Net~\cite{Bot2LNet2023} & N/A & 11.53 & 9.59 \\
 \hline
MCGaze~(Ours) & \textbf{12.96} & \textbf{10.74} & \textbf{10.02}\\
 \hline
\end{tabular}
\label{tab: all}
\vspace{-6mm}
\end{table}

% \begin{figure}[htb]
% \centerline{\includegraphics[width=0.6\columnwidth]{err_dis.png}}
% \caption{}
% \label{fig: err}
% \end{figure}

\subsection{Comparison with state-of-the-art methods}
% Recent work extracted video frames in Gaze360 where the subject's face could be recognized and used them for model evaluation. In our work, we use the Head-face-eye clues, which usually produces the best results when the subject's face can be detected, so we use a portion of the images in gaze360 where the face can be detected for training and testing.
% We follow Gaze360's dataset split and concern detectable faces, front$180^{\circ}$, and front facing for all three evaluations. 
The comparison with the state-of-the-art methods on the detectable face setting is shown in Table~\ref{tab: sub-dataset1}. We use the same training and testing set as the listed methods for a fair comparison. Essentially, our proposition outperforms the other methods in all the test cases, thus verifying its superiority.

Additionally, the comparison on the $360^{\circ}$ setting is shown in Table~\ref{tab: all}. Particularly, all models and methods in the table are trained using the entire Gaze360 dataset. 
In this more challenging setting, our approach still outperforms the state-of-the-art counterparts consistently. This indeed demonstrates the effectiveness and generality of our proposition. 

% In order to explore the relationship between the performance of our model for gaze prediction and the true gaze distribution, we plotted Fig. \ref{fig: err} to visualize this result. As shown in the figure, the better the model performs as both pitch and yaw angles approach $0^{\circ}$. And the performance of the model decreases overall when the line of sight is at an extreme angle.

Moreover, our model runs efficiently, achieving a processing speed of 70 FPS (inferencing within a video clip length of 7) on the Gaze360 dataset with a single RTX 3090. Our model has 83.09 M parameters and uses 28.01 GFLOPs.

\subsection{Ablation Study}
\textbf{Head-face-eye queries.} The effectiveness of concerning joint clues from the head, face, and eye in query form is verified in Table~\ref{tab: ablation1}. It can be observed that when all the 3 queries are used, the optimal performance can be acquired in all the test cases. This essentially reveals that, towards gaze estimation, the global clues from the head and face are complementary to local clues from the eye for leveraging performance. Besides, we notice that the feature degradation issue happens when there is only one head query. Specifically, for the Gaze360 benchmark, the input image is a human head image, so the network may learn more about the fixed head position and thus does not learn the gaze representation well. However, for the multi-clue case, the head query can provide useful global information as complementary and thus facilitate performance. Overall, adding more clues can facilitate gaze representation and boost performance consistently. 
\begin{table}[t]
\tiny
\vspace{-6mm}
\setlength{\abovecaptionskip}{0cm}  %段前
\setlength{\belowcaptionskip}{-0.2cm} %段后
\caption{Ablation Study.}
\centering

\setlength{\tabcolsep}{3pt}
\begin{tabular}{c|ccc}
\hline
Variants of MCGaze& Detectable faces & Front$180^{\circ}$ & Front facing\\
\hline
 MCGaze w/o face clue and eye clue  & 36.53 & 35.92  & 13.74  \\
 MCGaze w/o head clue and eye clue  & 10.62 & 10.42 & 8.33 \\
 MCGaze w/o head clue and face clue  & 10.87 & 10.60 & 8.12 \\
 MCGaze w/o eye clue  & 10.33 & 10.14 & 7.83 \\
 MCGaze w/o face clue   & 10.24 & 10.06 & 7.68 \\
 MCGaze w/o head clue  & 10.13 & 9.96 & 7.73 \\
 \hline
 MCGaze w/o spatial and temporal interaction  & 11.06 & 10.91 & 9.76\\
 MCGaze w/o temporal interaction  & 10.90 & 10.70 & 8.26\\
 MCGaze w/o spatial interaction  & 10.15 & 9.95 & 7.85\\
 \hline
 MCGaze w/o clue localization head  & 17.83 & 17.42 & 9.61\\
 \hline
MCGaze & \textbf{10.02} & \textbf{9.81} & \textbf{7.57} \\
 \hline
\end{tabular}
\label{tab: ablation1}
\vspace{-6mm}
\end{table}

\textbf{Spatial and temporal interaction in STQI.} The effectiveness of spatial and temporal interaction is also demonstrated in Table~\ref{tab: ablation1}. We can see that both spatial interaction and temporal interaction can facilitate performance consistently. When they are conducted jointly, the performance can be further enhanced. These indeed verify their effectiveness and the importance of head-face-eye spatial-temporal interaction context for video gaze characterization.

\textbf{Clue localization head in task-specific heads.} The effectiveness of clue localization head is shown in Table~\ref{tab: ablation1}. In MCGaze, we use this component to help query locate different clues, thereby boosting the performance of gaze estimation.
\vspace{-2mm}

\begin{figure}[h]
\centerline{\includegraphics[width=\columnwidth]{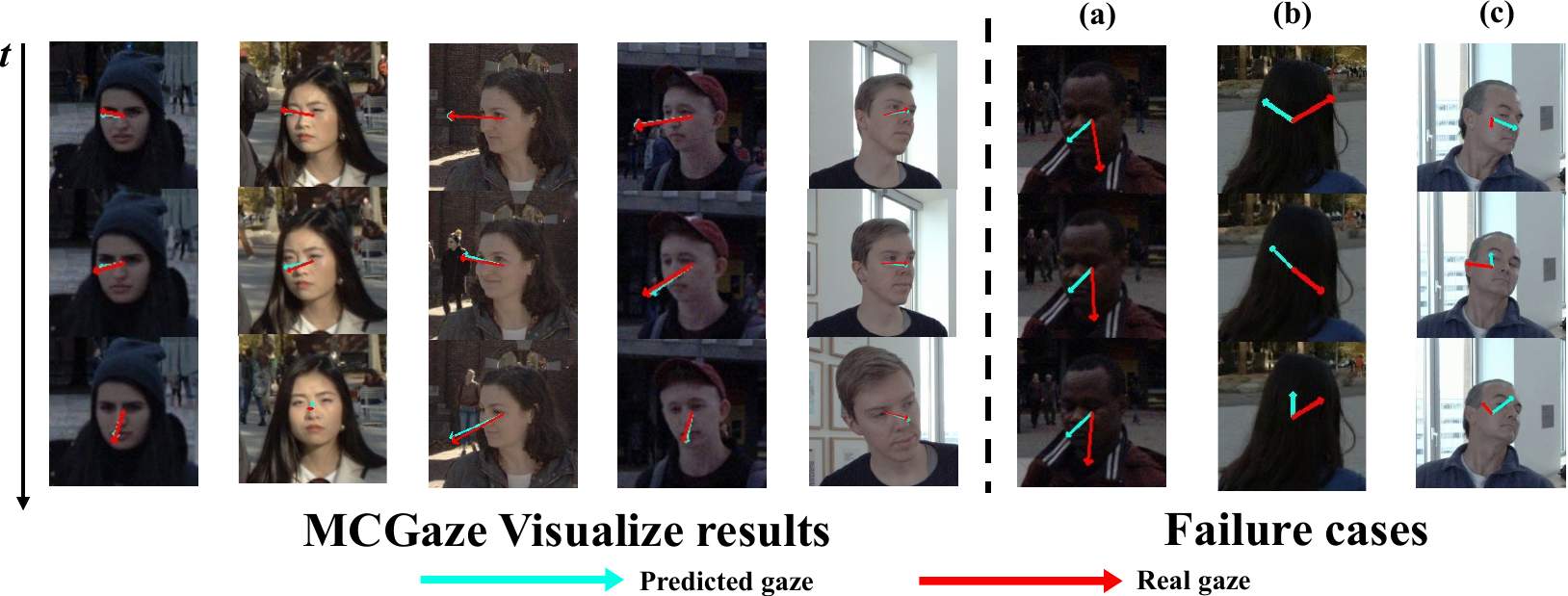}}
\vspace{-2mm}
\caption{Visualize result and failure case. Cyan and red arrows are prediction and GT respectively. Failure case: (a) Low imaging quality, (b) Invisible eyes, (c) Gaze and head directions in highly conflict.}
\label{fig:error}
\vspace{-6mm}
\end{figure}

\subsection{Qualitative analysis}
As shown in the left side of Fig.~\ref{fig:error}, MCGaze can produce excellent results under various environments, lighting, and gender. Some intuitive failure cases of our method are also given in the right part of Fig.~\ref{fig:error}. Specifically, our proposition cannot work well under some conditions: 
(a) Low imaging quality: The limited contextual information available from the images hinders the accuracy of the predicted gaze direction.
(b) Invisible eyes: The eye clues fail to capture local information from the eye region, leading to suboptimal predicted results.
(c) Gaze and head directions in highly conflict: The predicted gaze directions may be influenced by head directions.
% (1) low imaging quality. In this case, the clues are unable to acquire enough context from images, so our method cannot produce good predicted gaze direction.
% (2) Invisible eyes. Because the eyes are invisible, the eyes clue can not catch local information from eye region. Therefore, the predicted results are poor.
% (3) Extreme head direction. In this case, people's two eyes are not seen fully, So the predicted results may rely on the head direction more than eyes. 
% (1) eyes are invisible, (2) gaze and head are of highly conflict directions, and (3) low imaging quality.   

% Here we analyze the failure cases of the proposed method in Fig.~\ref{fig:error}. First, our method can not accurately predict the actual gaze direction when the whole eyes can not be visible due to the back of the face. However, we argue that it can still make a reasonable prediction based on the visible head clue. Second, the middle example indicates that the model will also fail when extreme gaze relative to the head pose occurs. Besides, our method may not work well under low-resolution conditions.

\section{CONCLUSIONS}
In this letter, we propose MCGaze to capture head-face-eye spatial-temporal interaction context well to facilitate video gaze characterization. 
% In a spatial-temporal query running way, our proposition can be trained with end-to-end optimization to seek optimal performance.
In an end-to-end learning way, our proposition can be trained to solve the tasks of clue localization and gaze estimation with joint optimization.
It achieves state-of-the-art performance on the challenging Gaze360 dataset with high running efficiency. 
However, our approach is tailored for individual subjects, and this presents a limitation. In the future, we will enhance this method to encompass multi-person scenarios and exploit richer spatial-temporal descriptive clues for video gaze estimation.

%\section*{References}
\bibliography{cite}

\begin{thebibliography}{10}
\providecommand{\url}[1]{#1}
\csname url@samestyle\endcsname
\providecommand{\newblock}{\relax}
\providecommand{\bibinfo}[2]{#2}
\providecommand{\BIBentrySTDinterwordspacing}{\spaceskip=0pt\relax}
\providecommand{\BIBentryALTinterwordstretchfactor}{4}
\providecommand{\BIBentryALTinterwordspacing}{\spaceskip=\fontdimen2\font plus
\BIBentryALTinterwordstretchfactor\fontdimen3\font minus
  \fontdimen4\font\relax}
\providecommand{\BIBforeignlanguage}[2]{{%
\expandafter\ifx\csname l@#1\endcsname\relax
\typeout{** WARNING: IEEEtran.bst: No hyphenation pattern has been}%
\typeout{** loaded for the language `#1'. Using the pattern for}%
\typeout{** the default language instead.}%
\else
\language=\csname l@#1\endcsname
\fi
#2}}
\providecommand{\BIBdecl}{\relax}
\BIBdecl

\bibitem{henderson2003human}
J.~M. Henderson, ``Human gaze control during real-world scene perception,''
  \emph{Trends in Cognitive Sciences}, vol.~7, no.~11, pp. 498--504, 2003.

\bibitem{SPL_gaze_object_segment}
R.~Shi, N.~K. Ngan, and H.~Li, ``Gaze-based object segmentation,'' \emph{IEEE
  Signal Processing Letters}, vol.~24, no.~10, pp. 1493--1497, 2017.

\bibitem{fan2018inferring}
L.~Fan, Y.~Chen, P.~Wei, W.~Wang, and S.-C. Zhu, ``Inferring shared attention
  in social scene videos,'' in \emph{Proceedings of the IEEE conference on
  computer vision and pattern recognition}, 2018, pp. 6460--6468.

\bibitem{fan2019understanding}
L.~Fan, W.~Wang, S.~Huang, X.~Tang, and S.-C. Zhu, ``Understanding human gaze
  communication by spatio-temporal graph reasoning,'' in \emph{Proceedings of
  the IEEE/CVF International Conference on Computer Vision}, 2019, pp.
  5724--5733.

\bibitem{emery2000eyes}
N.~J. Emery, ``The eyes have it: the neuroethology, function and evolution of
  social gaze,'' \emph{Neuroscience \& Biobehavioral Reviews}, vol.~24, no.~6,
  pp. 581--604, 2000.

\bibitem{zhang2019evaluation}
X.~Zhang, Y.~Sugano, and A.~Bulling, ``Evaluation of appearance-based methods
  and implications for gaze-based applications,'' in \emph{Proceedings of the
  CHI Conference on Human Factors in Computing Systems}, 2019, pp. 1--13.

\bibitem{zhang2015appearance}
X.~Zhang, Y.~Sugano, M.~Fritz, and A.~Bulling, ``Appearance-based gaze
  estimation in the wild,'' in \emph{Proceedings of the IEEE Conference on
  Computer Vision and Pattern Recognition}, 2015, pp. 4511--4520.

\bibitem{2eye_gaze}
Y.~Cheng, X.~Zhang, F.~Lu, and Y.~Sato, ``Gaze estimation by exploring two-eye
  asymmetry,'' \emph{IEEE Transactions on Image Processing}, vol.~29, pp.
  5259--5272, 2020.

\bibitem{nonaka2022dynamic}
S.~Nonaka, S.~Nobuhara, and K.~Nishino, ``Dynamic 3d gaze from afar: Deep gaze
  estimation from temporal eye-head-body coordination,'' in \emph{Proceedings
  of the IEEE/CVF Conference on Computer Vision and Pattern Recognition}, 2022,
  pp. 2192--2201.

\bibitem{bao2021adaptive}
Y.~Bao, Y.~Cheng, Y.~Liu, and F.~Lu, ``Adaptive feature fusion network for gaze
  tracking in mobile tablets,'' in \emph{Proceedings of the International
  Conference on Pattern Recognition}.\hskip 1em plus 0.5em minus 0.4em\relax
  IEEE, 2021, pp. 9936--9943.

\bibitem{cheng2020coarse}
Y.~Cheng, S.~Huang, F.~Wang, C.~Qian, and F.~Lu, ``A coarse-to-fine adaptive
  network for appearance-based gaze estimation,'' in \emph{Proceedings of the
  AAAI Conference on Artificial Intelligence}, vol.~34, no.~07, 2020, pp.
  10\,623--10\,630.

\bibitem{tip_gaze}
J.~Bao, B.~Liu, and J.~Yu, ``An individual-difference-aware model for
  cross-person gaze estimation,'' \emph{IEEE Transactions on Image Processing},
  vol.~31, pp. 3322--3333, 2022.

\bibitem{mtcnn}
K.~Zhang, Z.~Zhang, Z.~Li, and Y.~Qiao, ``Joint face detection and alignment
  using multitask cascaded convolutional networks,'' \emph{IEEE Signal
  Processing Letters}, vol.~23, no.~10, pp. 1499--1503, 2016.

\bibitem{SPL_Facial_Landmark}
Z.-H. Feng, J.~Kittler, and X.-J. Wu, ``Mining hard augmented samples for
  robust facial landmark localization with cnns,'' \emph{IEEE Signal Processing
  Letters}, vol.~26, no.~3, pp. 450--454, 2019.

\bibitem{faciallandmark_tip_23}
J.~Wan, J.~Liu, J.~Zhou, Z.~Lai, L.~Shen, H.~Sun, P.~Xiong, and W.~Min,
  ``Precise facial landmark detection by reference heatmap transformer,''
  \emph{IEEE Transactions on Image Processing}, vol.~32, pp. 1966--1977, 2023.

\bibitem{kellnhofer2019gaze360}
P.~Kellnhofer, A.~Recasens, S.~Stent, W.~Matusik, and A.~Torralba, ``Gaze360:
  Physically unconstrained gaze estimation in the wild,'' in \emph{Proceedings
  of the IEEE/CVF International Conference on Computer Vision}, 2019, pp.
  6912--6921.

\bibitem{sparse}
P.~Sun, R.~Zhang, Y.~Jiang, T.~Kong, C.~Xu, W.~Zhan, M.~Tomizuka, L.~Li,
  Z.~Yuan, C.~Wang \emph{et~al.}, ``Sparse r-cnn: End-to-end object detection
  with learnable proposals,'' in \emph{Proceedings of the IEEE/CVF Conference
  on Computer Vision and Pattern Recognition}, 2021, pp. 14\,454--14\,463.

\bibitem{Tevit}
S.~Yang, X.~Wang, Y.~Li, Y.~Fang, J.~Fang, W.~Liu, X.~Zhao, and Y.~Shan,
  ``Temporally efficient vision transformer for video instance segmentation,''
  in \emph{Proceedings of the IEEE/CVF Conference on Computer Vision and
  Pattern Recognition}, 2022, pp. 2885--2895.

\bibitem{instblink}
W.~Zeng, Y.~Xiao, S.~Wei, J.~Gan, X.~Zhang, Z.~Cao, Z.~Fang, and J.~T. Zhou,
  ``Real-time multi-person eyeblink detection in the wild for untrimmed
  video,'' in \emph{Proceedings of the IEEE/CVF Conference on Computer Vision
  and Pattern Recognition}, 2023, pp. 13\,854--13\,863.

\bibitem{garg2023multiscaled}
M.~Garg, D.~Ghosh, and P.~M. Pradhan, ``Multiscaled multi-head attention-based
  video transformer network for hand gesture recognition,'' \emph{IEEE Signal
  Processing Letters}, vol.~30, pp. 80--84, 2023.

\bibitem{spl_query_specific}
W.~Fu, L.~Zhou, and J.~Chen, ``Query-specific embedding co-adaptation improve
  few-shot image classification,'' \emph{IEEE Signal Processing Letters}, pp.
  1--5, 2023.

\bibitem{2023query_video}
S.~Huo, Y.~Zhou, R.~Wang, W.~Xiang, and S.-Y. Kung, ``Semantic relevance
  learning for video-query based video moment retrieval,'' \emph{IEEE
  Transactions on Multimedia}, 2023.

\bibitem{local2023}
Y.~Xiao, Q.~Yuan, K.~Jiang, X.~Jin, J.~He, L.~Zhang, and C.-w. Lin,
  ``Local-global temporal difference learning for satellite video
  super-resolution,'' \emph{arXiv preprint arXiv:2304.04421}, 2023.

\bibitem{local2020}
J.~Mun, M.~Cho, and B.~Han, ``Local-global video-text interactions for temporal
  grounding,'' in \emph{Proceedings of the IEEE/CVF Conference on Computer
  Vision and Pattern Recognition}, 2020, pp. 10\,810--10\,819.

\bibitem{liang2023local}
C.~Liang, W.~Wang, T.~Zhou, J.~Miao, Y.~Luo, and Y.~Yang, ``Local-global
  context aware transformer for language-guided video segmentation,''
  \emph{IEEE Transactions on Pattern Analysis and Machine Intelligence}, 2023.

\bibitem{cycmunet2023}
M.~Hu, K.~Jiang, Z.~Wang, X.~Bai, and R.~Hu, ``Cycmunet+: Cycle-projected
  mutual learning for spatial-temporal video super-resolution,'' \emph{IEEE
  Transactions on Pattern Analysis and Machine Intelligence}, 2023.

\bibitem{attention}
A.~Vaswani, N.~Shazeer, N.~Parmar, J.~Uszkoreit, L.~Jones, A.~N. Gomez,
  {\L}.~Kaiser, and I.~Polosukhin, ``Attention is all you need,'' \emph{In
  Proceedings of the Conference on Neural Information Processing Systems}, pp.
  5998--6008.

\bibitem{maskrcnn}
K.~He, G.~Gkioxari, P.~Doll{\'a}r, and R.~Girshick, ``Mask r-cnn,'' in
  \emph{Proceedings of the IEEE International Conference on Computer Vision},
  2017, pp. 2961--2969.

\bibitem{focalloss}
T.-Y. Lin, P.~Goyal, R.~Girshick, K.~He, and P.~Doll{\'a}r, ``Focal loss for
  dense object detection,'' in \emph{Proceedings of the IEEE International
  Conference on Computer Vision}, 2017, pp. 2980--2988.

\bibitem{Giou}
H.~Rezatofighi, N.~Tsoi, J.~Gwak, A.~Sadeghian, I.~Reid, and S.~Savarese,
  ``Generalized intersection over union: A metric and a loss for bounding box
  regression,'' in \emph{Proceedings of the IEEE/CVF Conference on Computer
  Vision and Pattern Recognition}, 2019, pp. 658--666.

\bibitem{cheng2022gaze}
Y.~Cheng and F.~Lu, ``Gaze estimation using transformer,'' in \emph{Proceedings
  of the International Conference on Pattern Recognition}.\hskip 1em plus 0.5em
  minus 0.4em\relax IEEE, 2022, pp. 3341--3347.

\bibitem{abdelrahman2022l2cs}
A.~A. Abdelrahman, T.~Hempel, A.~Khalifa, and A.~Al-Hamadi, ``L2cs-net:
  Fine-grained gaze estimation in unconstrained environments,'' \emph{arXiv
  preprint arXiv:2203.03339}, 2022.

\bibitem{yan2023gaze}
C.~Yan, W.~Pan, C.~Xu, S.~Dai, and X.~Li, ``Gaze estimation via strip pooling
  and multi-criss-cross attention networks,'' \emph{Applied Sciences}, vol.~13,
  no.~10, p. 5901, 2023.

\bibitem{kothari2021weakly}
R.~Kothari, S.~De~Mello, U.~Iqbal, W.~Byeon, S.~Park, and J.~Kautz,
  ``Weakly-supervised physically unconstrained gaze estimation,'' in
  \emph{Proceedings of the IEEE/CVF Conference on Computer Vision and Pattern
  Recognition}, 2021, pp. 9980--9989.

\bibitem{he2016deep}
K.~He, X.~Zhang, S.~Ren, and J.~Sun, ``Deep residual learning for image
  recognition,'' in \emph{Proceedings of the IEEE Conference on Computer Vision
  and Pattern Recognition}, 2016, pp. 770--778.

\bibitem{lin2017feature}
T.-Y. Lin, P.~Doll{\'a}r, R.~Girshick, K.~He, B.~Hariharan, and S.~Belongie,
  ``Feature pyramid networks for object detection,'' in \emph{Proceedings of
  the IEEE Conference on Computer Vision and Pattern Recognition}, 2017, pp.
  2117--2125.

\bibitem{deng2009imagenet}
J.~Deng, W.~Dong, R.~Socher, L.-J. Li, K.~Li, and L.~Fei-Fei, ``Imagenet: A
  large-scale hierarchical image database,'' in \emph{Proceedings of the IEEE
  Conference on Computer Vision and Pattern Recognition}, 2009, pp. 248--255.

\bibitem{adam}
I.~Loshchilov and F.~Hutter, ``Decoupled weight decay regularization,''
  \emph{arXiv preprint arXiv:1711.05101}, 2017.

\bibitem{zhang2017s}
X.~Zhang, Y.~Sugano, M.~Fritz, and A.~Bulling, ``It's written all over your
  face: Full-face appearance-based gaze estimation,'' in \emph{Proceedings of
  the IEEE Conference on Computer Vision and Pattern Recognition workshops},
  2017, pp. 51--60.

\bibitem{fischer2018rt}
T.~Fischer, H.~J. Chang, and Y.~Demiris, ``Rt-gene: Real-time eye gaze
  estimation in natural environments,'' in \emph{Proceedings of the European
  Conference on Computer Vision}, 2018, pp. 334--352.

\bibitem{chen2019appearance}
Z.~Chen and B.~E. Shi, ``Appearance-based gaze estimation using
  dilated-convolutions,'' in \emph{Proceedings of the Asian Conference on
  Computer Vision}.\hskip 1em plus 0.5em minus 0.4em\relax Springer, 2019, pp.
  309--324.

\bibitem{CADSE}
J.~O~Oh, H.~J. Chang, and S.-I. Choi, ``Self-attention with convolution and
  deconvolution for efficient eye gaze estimation from a full face image,'' in
  \emph{Proceedings of the IEEE/CVF Conference on Computer Vision and Pattern
  Recognition}, 2022, pp. 4992--5000.

\bibitem{NASgaze2023}
V.~Nagpure and K.~Okuma, ``Searching efficient neural architecture with
  multi-resolution fusion transformer for appearance-based gaze estimation,''
  in \emph{Proceedings of the IEEE/CVF Winter Conference on Applications of
  Computer Vision}, 2023, pp. 890--899.

\end{thebibliography}
\bibliographystyle{IEEEtran}
\end{document}